%% file: main.tex
\documentclass{article}
\usepackage[T1]{fontenc}
\usepackage{microtype}
\usepackage{graphicx}
\usepackage{subcaption}
\usepackage{booktabs}
\usepackage{multirow}
\usepackage{xcolor}
\usepackage{hyperref}
\usepackage[preprint]{icml2027}
\usepackage{amsmath}
\usepackage{amssymb}
\usepackage{mathtools}
\usepackage{amsthm}
\usepackage[capitalize,noabbrev]{cleveref}
\input{numbers}

\newcommand{\recov}{\mathrm{recovery}}

\icmltitlerunning{\bench{}: A Benchmark for Budgeted Video-Generation Routing}

\begin{document}

\twocolumn[
  \icmltitle{\bench{}: A Benchmark for Budgeted Video-Generation Routing}

  \begin{icmlauthorlist}
    \icmlauthor{Dayu Wang}{ntu,baidu}
  \end{icmlauthorlist}

  \icmlaffiliation{ntu}{Nanyang Technological University, Singapore}
  \icmlaffiliation{baidu}{Baidu Inc., Beijing, China}
  \icmlcorrespondingauthor{Dayu Wang}{dayu001@e.ntu.edu.sg}

  \icmlkeywords{video generation, model routing, cold start, multi-objective optimization}

  \vskip 0.3in
]

\printAffiliationsAndNotice{}

\begin{abstract}
Text-to-video generators differ by two orders of magnitude in price, and a cheap configuration of a strong model is
often good enough for a given prompt. Routing prompts across such \emph{arms} (model $\times$ sampling configuration)
should buy quality at fixed budget, but there is no dataset on which to study it: public video-preference sets contain
only closed models at a single configuration each. This paper makes three contributions.

\textbf{A benchmark.} \bench{} provides \nArms{} arms from \nFamilies{} open families spanning \costSpan$\times$ in
measured cost (\$\costMin--\$\costMax{} per clip), \nVideos{} generated clips (\gpuHours{} A800-hours) over
\nFullPrompt{} grouped prompts, a quality target frozen after calibration against human pairwise preferences
(\calibCV\% CV accuracy versus \calibBest\% for the best single scorer), per-clip scores on \nDims{} identifiable
dimensions, an aligned multi-seed noise panel, an encrypted test split and a sealed newest family. The evaluator compares
a router against the best \emph{prompt-agnostic} mixture. We also find that generating every prompt with the same seed
index, as is common, lets a router learn seed-specific artefacts: under independent per-prompt seeds the routing gain is
\gainIndepSeed{} instead of \gainHistSeed{}, and we release that protocol.

\textbf{An algorithm.} \algo{} (\emph{one probe, many utilities}) labels a new arm on $K$ probe prompts, predicts its
\nDims{} quality dimensions once, and serves any scenario utility by scalarising afterwards. Over \nW{} utility
directions it Pareto-dominates the prompt-agnostic baseline for \domMulti\% of directions, matching per-scenario routers
(\domDirect\%) without retraining per scenario.

\textbf{A finding: sample before you upgrade.} Seed-to-seed variation of one (prompt, arm) pair ($\sigma=\seedSdPanel$)
exceeds the gain of choosing the best arm for every prompt with hindsight (\oracleChoice). Drawing four samples from a
1.3B model and keeping the one a 2B reward model prefers costs \$\weakFourCost{} and beats one sample of a 14B
mixture-of-experts model at \$\strongOneCost{} ($\weakFour$ versus $\strongOne$ on the frozen target). Letting the
router choose the arm \emph{and} the number of samples, trained on best-of-$k$ targets, raises the gain over the
prompt-agnostic baseline from $+\rtRone$ to $+\rtMtwo$, while drafting at very low step counts to pick a seed does not pay.
Data, evaluator, method code and a worked example: \url{https://github.com/dayuwang401/VidRouteBench}.
\end{abstract}

\section{Introduction}
\label{sec:intro}

A production video service does not call one generator. Prices per five-second clip in our own measurements span
\costSpan$\times$, from \$\costMin{} for a distilled few-step model at reduced resolution to \$\costMax{} for a 14B
mixture-of-experts model, and the ordering of two arms flips across prompts. The natural response is to route: predict
quality per (prompt, arm), then pick $\arg\max_a(\hat q_{p,a}-\lambda c_a)$ and sweep $\lambda$ to trace a
cost--quality curve. The LLM community has converged on this formulation \citep{hu2024routerbench,ong2024routellm},
including its multi-objective reading \citep{fernandez2026radar}.

Two problems block deployment. First, \textbf{cold start}: a new generator arrives with no labels, and labelling it
densely is exactly the cost routing was supposed to avoid. Adaptive-testing ideas from LLM routing
\citep{polo2024tinybenchmarks,song2025irtrouter,fernandez2026radar} suggest probing a small, well-chosen set of prompts.
Second, \textbf{one utility is not enough}: an advertisement wants sharpness, a teaching clip wants text alignment, a
motion-heavy shot wants temporal quality. A single frozen scalar quality collapses these into one number and can route
every scenario to the same arm.

\begin{figure*}[!t]
  \centering
  \includegraphics[width=\textwidth]{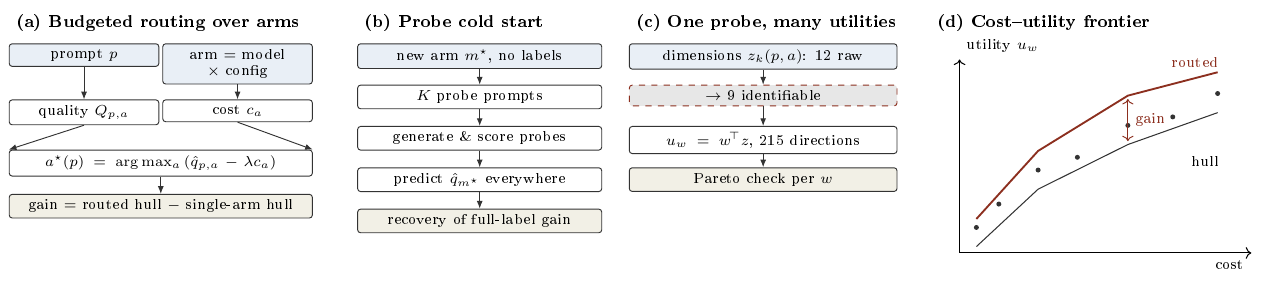}
  \caption{\textbf{Setup and findings.} (a) An arm is a model $\times$ sampling configuration; a router picks one
  arm per prompt under a budget and is scored against the upper hull of the \nArms{} single arms, the best policy that
  ignores the prompt. (b) A new arm is labelled on only $K\in\{10,20,50\}$ probe prompts, at probe cost
  $K\cdot c_{m^\star}$; recovery is the share of the full-label gain it recovers. (c) Scenarios are utility
  directions $w$ over the \nDims{} identifiable dimensions that remain after removing three collinear ones, and
  \algo{} predicts all of them from one probe set. (d) The reported quantity: vertical distance between the routed
  frontier and the prompt-agnostic hull, averaged over eight budgets.}
  \label{fig:overview}
\end{figure*}

We study both on self-generated data, because public video-preference sets contain only closed models at one
configuration each. Our setup (\cref{fig:overview}) treats an arm as a model \emph{plus} its sampling configuration,
which is what creates cheap-but-sometimes-sufficient options; this choice turns out to be the reason specialisation
appears at all.

\textbf{Contributions.}
\begin{enumerate}
\item \textbf{\bench{}, a benchmark for video-generation routing.} \nArms{} arms over \nFamilies{} open families with
\emph{configuration} arms (same model, fewer steps or lower resolution), measured cost spanning \costSpan$\times$,
\nVideos{} clips, \nFullPrompt{} semantically grouped prompts, a frozen human-calibrated quality target plus \nDims{}
identifiable per-dimension scores, an aligned multi-seed panel, an encrypted test split, a sealed newest family, and a
frozen evaluator whose baseline is the best prompt-agnostic mixture. We establish that the routing problem is
non-trivial (\cref{sec:gate}), provide the first baseline suite (\cref{sec:router,sec:cold}), and show that a
fixed seed index shared across prompts inflates routing gains by about 45\%, which motivates the independent-seed
protocol we release (\cref{sec:samples}).
\item \textbf{\algo{}, a probe algorithm that serves many utilities at once.} Rather than fitting one router per
scenario, \algo{} predicts the \nDims{} quality dimensions of a new arm from a single probe set and scalarises at
serving time, so a new scenario costs nothing. It Pareto-dominates the prompt-agnostic baseline on \domMulti\% of
\nW{} utility directions, on par with per-scenario routers (\domDirect\%) that must be retrained for every scenario
(\cref{sec:pareto}); the probe \emph{mean} is harmful, per-dimension profiling overfits the dev split, and Chebyshev
scalarisation buys nothing on a convex front.
\item \textbf{Weak models sampled several times can beat strong models sampled once.} Because seed-to-seed variation
($\sigma=\seedSdPanel$) is as large as the gap between arms, best-of-4 with a cheap reward-model judge adds
$+\bokLo$ to $+\bokHi$ to every arm, more than perfect per-prompt model choice (\oracleChoice). At matched cost, four
samples of Wan2.1-1.3B beat one sample of Wan2.2-A14B, and the sampling frontier saves \sampSave\% of cost at equal
quality on average. We extend routing to the action space (arm, number of samples): trained on best-of-$k$ targets, it
reaches $+\rtMtwo$ over the prompt-agnostic baseline against $+\rtRone$ for one-sample routing and $+\rtBO$ for
prompt-agnostic resampling, at every group size we tested (\cref{sec:samples}).
\end{enumerate}

\section{Related Work}
\label{sec:related}

\paragraph{Benchmarks for routing.} On the language side, RouterBench \citep{hu2024routerbench} supplies a
model$\times$query response matrix with prices so that routers can be compared offline, and RouteLLM
\citep{ong2024routellm} releases preference data for the two-model case, derived from Chatbot Arena votes
\citep{chiang2024arena}; both fix one configuration per model and measure cost in API tokens. Video generation has
benchmarks for \emph{quality} but not for \emph{routing}: VBench \citep{huang2024vbench} and VBench-2.0
\citep{zheng2025vbench2} score models on per-dimension suites, EvalCrafter \citep{liu2024evalcrafter} aggregates
reference-based and reference-free metrics, VideoGen-RewardBench \citep{liu2025videoalign} collects human pairwise
preferences over closed models, and VideoScore \citep{he2024videoscore} and VideoScore2 \citep{he2025videoscore2}
train judges to imitate them. None of these varies the sampling configuration of a model, which is where most of the
price range lives, and none reports per-clip compute cost; corpus-level metrics such as FVD
\citep{unterthiner2019fvd} give no per-prompt signal at all. \bench{} is built to fill exactly that gap, and we keep
RewardBench as a human-labelled control.

\paragraph{Routing algorithms.} Query-level routing has been formulated as classification against a strong model
\citep{ong2024routellm,ding2024hybrid}, as matrix factorisation of a model$\times$query matrix
\citep{zhuang2024embedllm}, as reward-guided expert selection \citep{lu2024zooter}, as cluster-level transfer to
unseen models \citep{jitkrittum2025uniroute}, and as item-response theory with latent ability and difficulty
\citep{lord1980irt,song2025irtrouter,fernandez2026radar}. A parallel line answers the same cost question
sequentially: cascades call a cheap model first and escalate when confidence is low
\citep{chen2023frugalgpt,madaan2023automix}. Cascading needs a cheap per-sample verifier, which this domain lacks ---
judging a clip costs a reward-model pass over its frames, and the clip has already been paid for --- so we decide
once, before generation. We port the three recipes that need no extra supervision (probe mean, per-arm profiling,
probe-only regression) and find that their ranking inverts on \bench{}.

\paragraph{Cold start and probe selection.} tinyBenchmarks \citep{polo2024tinybenchmarks} shows that a few anchor
items estimate a model's ability, and RADAR \citep{fernandez2026radar} adds adaptive testing so a new reasoning model
joins the router after a short test. These methods estimate a \emph{scalar} ability; \algo{} estimates a
\emph{vector} of quality dimensions from the same probe budget, which is what makes one probe set reusable across
scenarios.

\paragraph{Multi-objective and preference-conditioned decisions.} RADAR \citep{fernandez2026radar} states routing as
multi-objective optimisation over (quality, cost) and advocates Chebyshev scalarisation, which reaches concave parts
of a front that linear scalarisation misses \citep{miettinen1998mo,zhang2020chebyshev}. On the alignment side,
directional preference alignment \citep{wang2024dpa}, multi-objective DPO \citep{zhou2024modpo} and rewarded soups
\citep{rame2023rewardsoups} condition or interpolate \emph{one model} along a direction in reward space, which costs a
training run or a weight interpolation per direction. We keep the models fixed and move the trade-off into the routing
decision, so a new direction costs one dot product; we sweep directions over several quality dimensions across arms,
measure hypervolume \citep{zitzler1999hypervolume}, and report that Chebyshev's advantage does not materialise when
the front is convex.

\paragraph{Prompt representations and arms.} Our arms come from \nFamilies{} open families: CogVideoX
\citep{yang2025cogvideox}, Wan \citep{wan2025wan}, LTX-Video \citep{hacohen2024ltxvideo}, Kandinsky~5
\citep{arkhipkin2025kandinsky5} and, held out, HunyuanVideo~1.5 \citep{hunyuanvideo15}, all latent video diffusion
models in the lineage of \citet{blattmann2023videoldm} and all run through one diffusers pipeline
\citep{vonplaten2022diffusers}; the held-out family's predecessor \citep{kong2025hunyuanvideo} is not in the pool. On the prompt side, TF-IDF bag-of-words \citep{sparckjones1972idf} is the
standard cheap baseline, and our comparison against a sentence embedding \citep{zhang2025qwen3emb} in
\cref{sec:router} quantifies what semantic representation buys here, which is substantial exactly when labels are
scarce.

\section{\bench{}}
\label{sec:setup}

\paragraph{Problem setup.} Let $\mathcal{A}$ be a set of arms with per-clip costs $c_a>0$ and let $q_{p,a}$ be the
quality of arm $a$ on prompt $p$. A router is a map $\pi:p\mapsto a$ whose operating point is the pair
$\big(\mathbb{E}_p[c_{\pi(p)}],\,\mathbb{E}_p[q_{p,\pi(p)}]\big)$. Given a predictor $\hat q$, the family
$\pi_\lambda(p)=\arg\max_a(\hat q_{p,a}-\lambda c_a)$ traces a cost--quality curve as $\lambda$ sweeps from $0$ to
$\infty$, and mixing two values of $\lambda$ reaches every point on the upper convex hull of that curve. The reference
policy is the same construction with the prompt removed: the upper hull of the \nArms{} single-arm points, which is
the best operating curve available to a policy that cannot read the prompt. We report
\begin{equation}
  \mathrm{gain}=\tfrac{1}{8}\textstyle\sum_{b=1}^{8}\big[\mathcal{H}_\pi(b)-\mathcal{H}_0(b)\big],
  \label{eq:gain}
\end{equation}
with $\mathcal{H}_\pi,\mathcal{H}_0$ the routed and prompt-agnostic hulls at eight budgets spanning the attainable
range. A positive gain means per-prompt choice buys quality that no fixed mixture of arms can.

\paragraph{Arms.} An arm is a model with a fixed sampling configuration (\cref{tab:arms}). Configurations of the same
model matter: halving steps or lowering resolution moves an arm by $2$--$4\times$ in price, and whether that costs
quality is prompt-dependent. We measure cost as GPU-seconds on one A800 times \$1.5 per GPU-hour, taking the minimum
over clips because concurrent jobs inflate wall time.

\paragraph{Prompts.} \nPool{} prompts from VBench \citep{huang2024vbench}, VBench-2.0 \citep{zheng2025vbench2} and
TA-Hard, de-duplicated and grouped by cosine similarity above $0.92$; groups do not cross splits (train \nTrain{} /
dev \nDev{} / test \nTest{}). The test split is encrypted and untouched. We generated \nGenPrompt{} prompts for all
arms and \nFullPrompt{} for the cheaper ones.

\paragraph{Frozen quality target.} Every clip is scored by VideoReward \citep{liu2025videoalign} (VQ/MQ/TA) and by
SigLIP2-based \citep{tschannen2025siglip2} alignment, temporal and low-level statistics. On \calibVideos{} RewardBench
clips with \calibPairs{} non-tied human comparisons we fit $\mathrm{target}=\sum_i w_i z_i$ with $w_i\ge0$ by logistic
regression on score differences \citep{bradley1952bt}, grouped 5-fold CV: \calibCV$\pm$\calibCVsd\% accuracy versus
\calibBest\% for the best single scorer. Non-negativity keeps the target monotone in every component, so an arm cannot
gain by scoring worse on a dimension. The weights, standardisation and scorer versions are frozen and hashed before
any routing experiment; \cref{app:target} lists the fitted weights.

\paragraph{Evaluator.} For predicted quality $\hat q$ we sweep $\lambda$, take the upper convex hull of routing points
together with single-arm points, and subtract the single-arm hull, averaging over eight budgets. Hulls allow
proportional mixing, so the baseline is the strongest prompt-agnostic policy. For cold start, each non-held-out arm in
turn becomes the new arm $m^\star$, labelled only on $K\in\{10,20,50\}$ probe prompts chosen by the method, and we
report $\recov=(\mathrm{gain}-\mathrm{gain_{excl}})/(\mathrm{gain_{full}}-\mathrm{gain_{excl}})$ together with probe
cost $K\cdot c_{m^\star}$. The evaluator, including the target, is frozen; methods are separate files.

\begin{table}[t]
  \centering
  \small
  \caption{The \nArms{} arms, sorted by measured cost. Mean $Q$ is the frozen target averaged over prompts.}
  \label{tab:arms}
  \resizebox{\columnwidth}{!}{\input{tables/tab_arms}}
\end{table}

\section{Does per-prompt specialisation exist?}
\label{sec:gate}

Our prior from LLM-style arms was negative, so we gate the whole study on this question
(\cref{tab:gate}). The oracle gap is \oracleGap{} in target units. Selecting the best arm on one seed and scoring on
another retains \gapCross{} of the \gapWithin{} same-seed gap, so \gapReal\% of the gap is reproducible rather than
seed noise, which is large here (seed s.d.\ \seedSd{}). A two-way analysis with seed replication gives interaction
$F(447,600)=\interF$ ($p=\interP$), with variance components \interVar{} for interaction against \noiseVar{} for
seeds. Finally, regressing each arm's interaction residual on prompt embeddings gives out-of-fold $R^2$ between
\interRtwoLo{} and \interRtwoHi{} (mean \interRtwoMean{}); full-label routing gains \fullGain{} against
\shufGain{} for shuffled labels.

The same pipeline on RewardBench's twelve closed models finds almost no predictable specialisation. We attribute the
difference to configuration arms: the question "which prompts tolerate the cheap setting" is itself a strong
prompt-level signal, and it does not exist in a pool of one-configuration closed models.

\begin{table}[t]
  \centering
  \small
  \caption{Gate results. All numbers on self-generated data, dev split where applicable.}
  \label{tab:gate}
  \resizebox{\columnwidth}{!}{\input{tables/tab_gate}}
\end{table}

\section{Baselines: what the prompt representation buys}
\label{sec:router}

Before any cold start, we fix the router that predicts quality for arms that \emph{do} have labels, since every
cold-start number is measured against it. \cref{tab:router} compares a bag-of-words representation (TF-IDF over
1--2 grams, computed from the prompt text inside the method so that no shared feature file is touched) against
Qwen3-Embedding-0.6B, both with ridge regression.

With the full 5-fold pool the two are closer than expected: bag-of-words reaches \fgBow{} against \fgEmb{} for the
embedding, a 17\% relative loss. In the low-data regime that matters for a new deployment (train $\rightarrow$ dev,
\nTrain{} prompts) the gap becomes decisive: \fgBowDev{} against \fgEmbDev{}, i.e.\ bag-of-words gives up almost all
of the routing gain and lands near the shuffled-label control. Word overlap is enough to separate arms once thousands
of labelled prompts exist, and nearly useless before that.

\begin{table}[t]
  \centering
  \small
  \caption{Router baselines on \bench{}. Gain is the routed frontier minus the single-arm hull, averaged over eight
  budgets; higher is better.}
  \label{tab:router}
  \resizebox{\columnwidth}{!}{\input{tables/tab_router}}
\end{table}

\section{Cold start: which classical recipes transfer}
\label{sec:cold}

We now hold the router of \cref{sec:router} fixed (embedding ridge) and vary only what the probe budget buys
(\cref{tab:cold}). \textsf{mean} credits the new arm with its probe average, \textsf{ridge}$_K$ fits
prompt$\rightarrow$quality on the $K$ probes alone, and \textsf{profile} regresses the new arm's probe quality on the
known arms' quality at the same prompts and transfers that map.

Only \textsf{profile} is positive under 5-fold CV ($\cvProfile{}$). \textsf{mean} is strongly negative
($\cvMean{}$): with a \costSpan$\times$ price span, an arm that is cheap and \emph{sometimes} good gets selected
everywhere once it is credited with a flat average, so the router is worse off than if the arm were unavailable. This
is the exact opposite of RewardBench-style arms, where \textsf{mean} is the strong baseline ($\rbMean{}$) and
profiling adds little. \textsf{ridge}$_K$ is also negative ($\cvRidge{}$): $K\le50$ prompts cannot fit a
prompt-conditioned model from scratch.

\paragraph{A variant we reject.} Since the dimensions carry different information, we tried \textsf{dim-profile},
which profiles each of the \nDims{} dimensions separately against the known arms and recombines them with a map fitted
on the probes. It is the best method on the dev split ($\dvDim{}$ against $\dvProfile{}$ for \textsf{profile}) and
\emph{worse} under 5-fold CV ($\cvDim{}$). We report it as a negative result and as a caution about the dev split:
with \nDev{} dev prompts and five usable arms, a $+0.04$ dev improvement is not evidence.

\paragraph{What the numbers cannot yet support.} Only five arms clear the $\mathrm{room}>0.003$ filter, the largest
contributing $0.039$ of attainable gain, so recovery denominators are small and the ratios swing widely. The ordering
(\textsf{profile} $>$ \textsf{dim-profile} $>$ \textsf{ridge}$_K$ $>$ \textsf{mean}) is stable across splits; the
magnitudes are not. Generation of more arms and seeds is ongoing.

\begin{table}[t]
  \centering
  \small
  \caption{Cold start on \bench{} with the router held fixed. 5-fold CV is the decision criterion; the dev column
  shows how misleading a single split is.}
  \label{tab:cold}
  \resizebox{\columnwidth}{!}{\input{tables/tab_cold}}
\end{table}

\section{\algo{}: one probe set, many utilities}
\label{sec:pareto}

\paragraph{Why a scalar target is not enough.} A frozen scalar quality collapses sharpness, motion quality and text
alignment into one number, and on \bench{} the per-dimension winners differ: of \nDimsRaw{} raw dimensions, nine
distinct arms are best on at least one. An advertisement, a teaching clip and a motion-heavy shot are different
scenarios, and routing them with one scalar sends them to the same arm.

\paragraph{Identifiable dimensions.} Three of the \nDimsRaw{} raw dimensions are redundant \emph{by construction}:
dynamic degree is $1-$consistency, VideoReward's Overall is the sum of its three sub-scores, and minimum alignment
correlates $0.93$ with mean alignment. Removing them takes the correlation condition number from \condBefore{} to
\condAfter{}, leaving \nDims{} dimensions in which a weight vector is identifiable. We fit the standardisation on
train only and freeze it, and we keep the frozen scalar target as one direction in the grid so that results stay
comparable with \cref{sec:cold}.

\paragraph{The algorithm.} \algo{} takes the probe set of \cref{sec:cold} and (i) predicts all \nDims{} dimensions of
the new arm with one multi-output ridge, (ii) stores that prediction, and (iii) at serving time scalarises
$u_w=w^{\!\top}z$ for whatever direction $w$ the scenario asks for and routes with
$\arg\max_a(\hat u_w-\lambda c_a)$. The probe budget is paid once; a new scenario costs one dot product. The
alternative --- one router per scenario, trained on the scalarised target --- pays the probe budget and a training run
per scenario.

\paragraph{Evaluation.} We evaluate \nW{} directions: six named scenarios, \nDims{} single-dimension corners and
\nWdir{} Dirichlet($\mathbf{1}$) samples. Per direction we report the gain above the single-arm hull, hypervolume in
the normalised cost--utility plane, CPT(90\%), and whether the routed frontier dominates at all eight budgets.

\cref{tab:pareto} gives the result. Under oracle utilities the routed frontier dominates for \emph{all} \nW{}
directions ($\Delta$HV $+\hvOracle{}$, worst direction $+\gainOracleMin{}$), so the Pareto improvement is a property
of the arm pool and not of favourable weights. With predicted utilities \algo{} dominates on \domMulti\% of directions
at mean gain $+\gainMulti{}$, against \domDirect\% and $+\gainDirect{}$ for per-scenario routers: predicting the
dimensions once and scalarising later matches one router per scenario while training once. On the earlier 899-prompt
snapshot \algo{} was ahead (94\% versus 90\%); with all \nFullPrompt{} prompts the difference vanishes, so we claim
the cost advantage, not an accuracy advantage. The remaining distance to oracle
($\gainOracle{}$ versus $\gainMulti{}$) is headroom in utility prediction, which is where we expect the next gain.

\paragraph{Chebyshev does not help here.} RADAR \citep{fernandez2026radar} reports that Chebyshev scalarisation beats
linear scalarisation because it reaches concave parts of the Pareto front. On \bench{} it is worse
($+\gainCheb{}$, \domCheb\%). Our cost--utility front is essentially convex: with a \costSpan$\times$ price span and
monotone quality there is no concave region to reach, and the ideal point that Chebyshev needs must itself be
estimated from the probes, adding variance. We read this as a boundary condition on RADAR's recommendation.

\begin{table}[t]
  \centering
  \small
  \caption{\algo{} against alternatives over \nW{} utility directions (statistics over the \nWdir{} Dirichlet
  samples). Dominance is the fraction of directions where the routed frontier beats the prompt-agnostic hull at all
  eight budgets.}
  \label{tab:pareto}
  \resizebox{\columnwidth}{!}{\input{tables/tab_pareto}}
\end{table}

\section{Routing over samples: a weak model sampled $k$ times can beat a strong model sampled once}
\label{sec:samples}

\paragraph{Evaluation protocol.} \texttt{diffusers} pipelines seeded with a fixed index start every prompt from the same
initial latent, so ``seed 0 of arm $a$'' is a shared perturbation across prompts that a router can learn. Training on
seed~0 and evaluating on dev prompts generated with seed~0 gives a routing gain of \gainHistSeed{} over 16 arms and
405 dev prompts; evaluating the same router on dev clips generated with independent per-prompt seeds gives
\gainIndepSeed{} (cheapest 11 arms: $0.343\rightarrow0.132$). Every number in this section is computed on seeds that
differ from the training seed, and the release includes the independent-seed dev split.

\begin{figure*}[t]
  \centering
  \includegraphics[width=\textwidth]{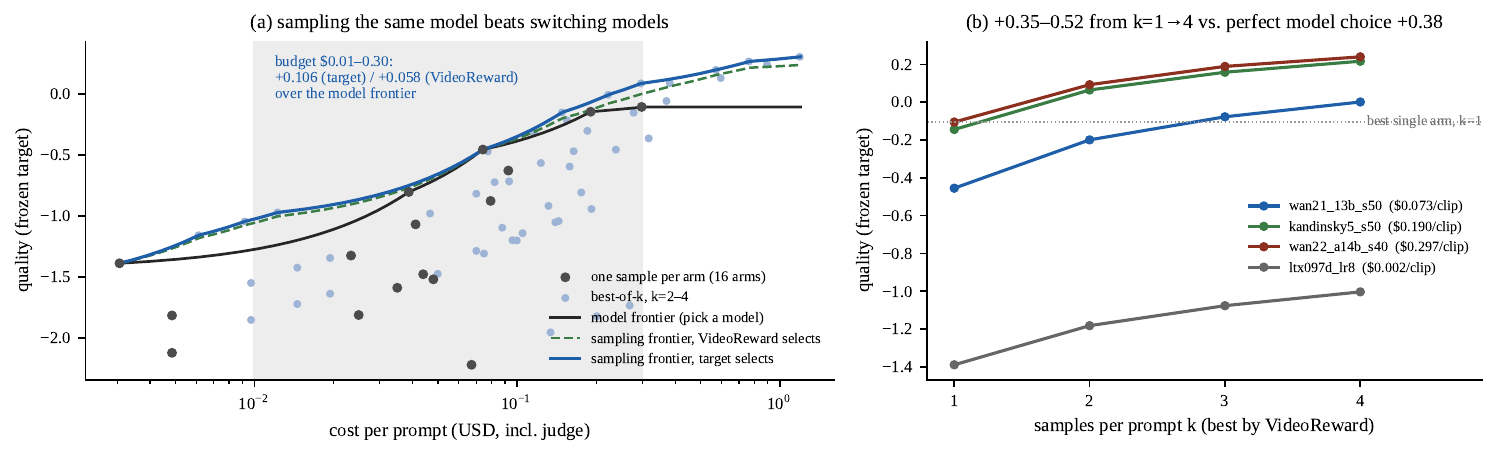}
  \caption{\textbf{Sampling versus switching models.} (a) Cost--quality points of every arm at $k{=}1$ (dark) and
  best-of-$k$, $k{=}2$--4 (light), with the one-sample model frontier and the sampling frontier when VideoReward
  (dashed) or the target itself (solid) selects. (b) Best-of-$k$ curves for four arms; the dotted line is the best
  single arm at $k{=}1$. \nPanelBok{} panel prompts, seeds 0/2/3/4.}
  \label{fig:samples}
\end{figure*}

\paragraph{Variation within an arm rivals variation across arms.} On \nPanelBok{} panel prompts with four aligned seeds,
the standard deviation of the frozen target across seeds of one (prompt, arm) pair is \seedSdPanel{}, the spread of arm
means is \armSpread{}, and choosing the best arm for every prompt with hindsight gains \oracleChoice{} over the best
single arm. Generating $k$ samples and keeping the one VideoReward Overall prefers (a 2B judge that costs about
\$0.0006 per clip), then scoring with the frozen target, adds $+\bokLo$ to $+\bokHi$ per arm at $k{=}4$
(\cref{fig:samples}b). The judge is not the target, so this is not the target selecting itself; with the target as its
own selector the gains are larger still.

\paragraph{Matched cost.} Four samples of Wan2.1-1.3B at 50 steps cost \$\weakFourCost{} and reach $\weakFour$; one
sample of Wan2.2-A14B costs \$\strongOneCost{} and reaches $\strongOne$. Over budgets \$0.01--0.30 the frontier of all
(arm, $k$) points lies $+\sampFront$ above the one-sample model frontier on average (\cref{fig:samples}a), and reaching a
given quality costs \sampSave\% less on average and \sampSaveTop\% less at the quality of the best single arm.

\paragraph{Routing over (arm, $k$).} We let the router pick an arm and a number of samples per prompt,
$\arg\max_{a,k}\hat u(p,a,k)-\lambda k(c_a+c_{\mathrm{judge}})$, and compare three ways of estimating the value of
$k$: a per-arm constant $g_k(a)$ estimated on training prompts (M1), a prompt-dependent spread
$\hat\sigma(p,a)\,\mathbb{E}[\max_k]$ regressed from prompt features (M3, ``sample more where the outcome is
uncertain''), and direct regression of best-of-$k$ targets for each (arm, $k$) (M2). \cref{tab:samples} reports \nDevBok{}
panel dev prompts whose four evaluation seeds all differ from the training seed. Resampling without reading the prompt
gains $+\rtBO$; one-sample routing gains $+\rtRone$; routing over (arm, $k$) gains $+\rtMone$, $+\rtMthree$ and
$+\rtMtwo$ for M1, M3 and M2. The two levers add up, and training on the quantity the policy actually delivers (the
best of $k$) is better than adding a constant. Prompt features predict per-pair spread only weakly (out-of-sample
correlation 0.27), which is why M3 sits between M1 and M2.

\begin{table}[t]
  \centering
  \small
  \caption{Routing over (arm, number of samples), budget shared by all \nDevBok{} panel dev prompts; win rate converts
  the gain to a predicted pairwise preference through the logistic link of the frozen target. Last column: share of
  random 5-prompt groups in which the policy is worse than prompt-agnostic resampling.}
  \label{tab:samples}
  \resizebox{\columnwidth}{!}{\input{tables/tab_samples}}
\end{table}

\paragraph{Group size.} When the budget must hold within groups of 5, 10, 20 or 40 prompts, the mean gain of M2 stays
between $+0.152$ and $+0.160$, but the share of groups where it loses to prompt-agnostic resampling falls from 40\% to
13\%. Resampling helps small groups slightly more ($+0.056$ over one-sample routing at size 5), because it gives the
knapsack a finer-grained way to spend the budget.

\paragraph{What does not work: drafting.} An appealing variant drafts several low-step samples, keeps the seed the
judge prefers and re-renders that seed at full steps. Low- and high-step samples of the same seed are correlated, but
after removing the prompt mean the seed effect correlates only 0.09--0.19 at 5--8 steps (0.3--0.5 at half the steps),
and the drafted seed is the full-step winner 28--36\% of the time against 25\% by chance. Every draft-then-refine policy
lies 0.03--0.38 below the plain best-of-$k$ frontier of the same family. The very-low-step configurations are still
useful as arms: each becomes the cheapest vertex of its family's frontier.

\section{Limitations}
\label{sec:limits}

The target is an automatic score calibrated to human preferences, not human judgement; its pairwise accuracy is
\calibCV\%, so conclusions are reported with the RewardBench human-label check beside them. Seed noise is large
(s.d.\ \seedSd{}), and only a subset of arms currently has three seeds. Recovery denominators are small, as discussed
in \cref{sec:cold}. VideoScore2 and VBench scorers are not yet deployed, so the dimension set is narrower than
intended. The newest family and the encrypted test split are deliberately untouched, so no held-out number is reported
here. Finally, all costs are our own GPU-seconds, not vendor prices. The best-of-$k$ results rest on the frozen target and
an automatic judge; a human check of selected versus random samples is pending, and in a latency-bound service the
$k$ samples must run in parallel, which multiplies peak GPU demand by $k$. The (arm, $k$) routing table uses
\nDevBok{} dev prompts. On the algorithm side, \algo{} is evaluated against
per-scenario ridge and against oracle utilities, not against a learned multi-objective router; the probe set itself is
still chosen at random, and direction-aware probe selection is the obvious next step.

\section{Conclusion}
\label{sec:concl}

\bench{} makes budgeted video-generation routing measurable: configuration arms create a \costSpan$\times$ price range
in which per-prompt choice provably pays, the quality target is calibrated to human preferences and then frozen, and
the evaluator compares against the best prompt-agnostic mixture rather than a single arm. On this data the classical
cold-start recipes reorder: profiling a new arm against known arms transfers, while crediting it with its probe
average is actively harmful, the reverse of what holds for one-configuration closed models.

\algo{} turns the probe budget into a vector rather than a scalar: predict the identifiable quality dimensions once,
scalarise at serving time, and a new scenario is free. It Pareto-dominates the prompt-agnostic baseline on
\domMulti\% of \nW{} utility directions, on par with per-scenario routers while training once. Two negative results
delimit the design space --- bag-of-words prompt features collapse in the low-data regime, and Chebyshev scalarisation
buys nothing on a convex front --- and both follow from properties of this domain rather than from implementation.

The largest lever, however, is not which model to call but how many times to call it. Seed-to-seed variation is as
large as the gap between models, so four samples of a small model, filtered by a 2B judge, beat one sample of the
largest model at the same price. A router whose actions are (arm, number of samples) and whose targets are best-of-$k$
qualities combines both levers; we expect sampling-aware routing, not finer model selection, to be where budgeted video
generation gains most.

\section*{Reproducibility}
The benchmark, the frozen target and its hash, the evaluator, all method files and the scripts that generate every
number and figure are released with the paper. Running order and raw logs are in the repository.
\textbf{Version note.} This version adds \cref{sec:samples}, the independent-seed protocol, and the full
\nFullPrompt{}-prompt multi-utility numbers (\cref{tab:pareto}; the earlier 899-prompt snapshot reported 94\% versus
90\%). The gate, router and cold-start tables (\cref{tab:gate,tab:router,tab:cold}) are unchanged and use the fixed-seed
protocol, so their routing gains should be read as upper bounds; \cref{sec:samples} gives independent-seed numbers.
The repository's \texttt{examples/} folder reproduces \cref{tab:samples} and the matched-cost comparison on CPU.

\bibliography{refs}
\bibliographystyle{icml2027}

\appendix
\onecolumn
\section{Benchmark construction details}
\label{app:bench}

\paragraph{Generation.} All clips are produced with one \texttt{diffusers} \citep{vonplaten2022diffusers} driver on
8$\times$A800-80GB, one worker process per GPU, \texttt{bfloat16}, no CPU offload unless the arm does not fit (the
14B mixture-of-experts arm runs alone on a GPU). Each worker claims prompts from a file-backed queue and appends a
completion record, so a crashed run resumes without regenerating. Negative prompts, guidance scales and schedulers are
the defaults shipped with each checkpoint; the only quantities we vary inside a family are the step count and the
output resolution, which is what defines a configuration arm. Clip length is the family default (49--121 frames,
5~seconds nominal).

\paragraph{Cost measurement.} Cost is wall-clock GPU-seconds for one clip, excluding model load, times \$1.5 per
GPU-hour. We take the \emph{minimum} over all clips of an arm rather than the median: with up to two generation
processes per GPU, concurrency inflates wall time by up to $2\times$ (for example \texttt{wan21\_13b\_s25} measures
92\,s alone and 187\,s under sharing), and the minimum is the only concurrency-free estimate available offline.

\paragraph{Prompt pool and splits.} \nPool{} prompts are pooled from VBench \citep{huang2024vbench}, VBench-2.0
\citep{zheng2025vbench2} and the hard text-alignment subset we call TA-Hard, lower-cased, de-duplicated exactly, then
embedded and single-link grouped at cosine similarity $0.92$ so that near-paraphrases stay together. Groups are
assigned to train/dev/test (\nTrain{}/\nDev{}/\nTest{}) so no group crosses a split, and the maximum cosine
similarity from dev and test back to train is $0.893$ and $0.910$. The test prompts ship only as an encrypted file
plus its SHA-256; the passphrase is held by a human and the file has never been decrypted in this work. Generation covers \nFullPrompt{} prompts for arms cheap enough to afford
them and \nGenPrompt{} for the rest, giving \nVideos{} scored clips (\gpuHours{} A800-hours) with seed~0 complete and
seeds~1--3 partial (21{,}703 / 1{,}797 / 1{,}116 / 2{,}312 clips).

\section{Scoring and the frozen target}
\label{app:target}

\paragraph{Scorers.} VideoReward \citep{liu2025videoalign} (\texttt{KwaiVGI/VideoReward}, Qwen2-VL-2B backbone,
\texttt{checkpoint-11352}) gives VQ, MQ, TA and Overall; it runs with \texttt{sdpa} attention and batch size~1 because
FlashAttention-2 is unavailable on this driver, and that setting is recorded in the frozen config. SigLIP2
\citep{tschannen2025siglip2} (\texttt{google/siglip2-so400m-patch16-384}, 16 sampled frames) gives
prompt--frame alignment (mean, min, std over frames) plus reference-free statistics: inter-frame motion, temporal
consistency, Laplacian sharpness, RMS contrast and colourfulness. Twelve raw dimensions in total.

\paragraph{Calibration.} Each dimension is oriented (so larger is better), standardised with train-split
$\mu,\sigma$, and combined as $\mathrm{target}=\sum_i w_i\,o_i\,(s_i-\mu_i)/\sigma_i$. The weights are fitted on
VideoGen-RewardBench: \calibVideos{} clips, \calibPairs{} non-tied human Overall comparisons, logistic regression on
score \emph{differences} with $w\ge0$ (L-BFGS-B with a box constraint). Grouped 5-fold CV by prompt gives
\calibCV$\pm$\calibCVsd\% pairwise accuracy against \calibBest\% for the best single scorer (VideoReward Overall) and
$52$--$60\%$ for every other individual dimension. Fitted weights (rounded): \texttt{vr\_ta} 0.600,
\texttt{vr\_overall} 0.461, \texttt{contrast} 0.201, \texttt{vr\_mq} 0.194, \texttt{consistency} 0.174,
\texttt{motion} 0.174, \texttt{vr\_vq} 0.130, \texttt{sharp} 0.103, \texttt{colorfulness} 0.095,
\texttt{align\_mean} 0.071, and $0$ for \texttt{align\_min} and \texttt{align\_std}. The whole config (weights,
$\mu$, $\sigma$, orientations, scorer repos and revisions) is serialised as \texttt{configs/target.json}, version
\texttt{target-v1-20261111}, SHA-256 \texttt{97f7c311\ldots3da}, and the evaluator refuses to run if the hash changes.

\section{Evaluator}
\label{app:eval}

For a quality matrix $\hat q\in\mathbb{R}^{P\times A}$ the evaluator sweeps $161$ values of $\lambda$ (zero plus a
log grid from $10^{-3}$ to $10^{2}$, rescaled by the cost range), computes the realised (cost, true-quality) point of
each $\pi_\lambda$, takes the upper convex hull of those points together with the \nArms{} single-arm points, and
reports \cref{eq:gain} at eight budgets evenly spaced strictly inside the attainable cost range. The prompt-agnostic
reference $\mathcal{H}_0$ is the upper hull of the single-arm points alone, which is exactly what budget-constrained
proportional mixing of arms can achieve without reading the prompt; comparing against the best single arm instead
would overstate every method.

For cold start each non-held-out arm $m^\star$ in turn is treated as new: its labels are hidden except on $K$ probe
prompts, $K\in\{10,20,50\}$, chosen by the method, and
\[
  \recov=\frac{\mathrm{gain}-\mathrm{gain_{excl}}}{\mathrm{gain_{full}}-\mathrm{gain_{excl}}},
\]
where $\mathrm{gain_{excl}}$ drops the arm from the pool and $\mathrm{gain_{full}}$ labels it densely. Arms with
$\mathrm{gain_{full}}-\mathrm{gain_{excl}}\le\mathrm{ROOM\_MIN}=0.003$ are excluded from the average because their
denominator is noise; five arms currently clear the filter, the largest contributing $0.039$. Probe cost is reported
separately as $K\cdot c_{m^\star}$. The evaluator, the target and the split assignment live in files that methods
import but cannot write.

\section{Methods and hyper-parameters}
\label{app:methods}

All routers are ridge regressions \citep{hoerl1970ridge} with the penalty chosen by generalised cross-validation over
$\alpha\in\{10^{-1},\ldots,10^{4}\}$ (11 log-spaced values), fitted per arm on centred targets.

\begin{itemize}
  \item \textsf{bow}: TF-IDF \citep{sparckjones1972idf} over 1--2 grams of the raw prompt, sublinear term frequency,
  \texttt{min\_df}=3, at most 4096 features, English stop-words. The vectoriser is fitted inside the method on the
  training prompts only, so no shared feature file is involved.
  \item \textsf{emb}: Qwen3-Embedding-0.6B \citep{zhang2025qwen3emb} sentence embeddings (1024-d), L2-normalised.
  \item \textsf{mean}: the new arm is credited with the mean of its $K$ probe scores on every prompt.
  \item \textsf{ridge}$_K$: a ridge router fitted on the $K$ probe rows of the new arm alone.
  \item \textsf{profile}: regress the new arm's probe quality on the known arms' quality at the same prompts
  (least squares with intercept, $A_{\text{known}}$ coefficients), then apply that map to all prompts.
  \item \textsf{dim-profile} (rejected): the same construction per quality dimension, recombined by a second least
  squares fitted on the probes.
  \item \algo{}: one multi-output ridge predicts all \nDims{} dimensions of the new arm; serving scalarises
  $u_w=w^\top z$ and routes with $\arg\max_a(\hat u_w-\lambda c_a)$.
\end{itemize}

Reported recovery numbers are 5-fold cross-validated over prompt groups; the dev column of \cref{tab:cold} is a
single split and is shown only to document how far a single split can mislead ($+0.25$ dev versus $-0.23$ CV for
\textsf{dim-profile}).

\section{Utility basis and scenarios}
\label{app:utility}

The \nDimsRaw{} raw dimensions contain exact redundancies: \texttt{motion} $=1-$\texttt{consistency} (correlation
$-1.00$ by construction), \texttt{vr\_overall} $=$ \texttt{vr\_vq}$+$\texttt{vr\_mq}$+$\texttt{vr\_ta}, and
\texttt{align\_min} correlates $0.93$ with \texttt{align\_mean}. Dropping the three leaves the basis
\{\texttt{vr\_vq}, \texttt{vr\_mq}, \texttt{vr\_ta}, \texttt{align\_mean}, \texttt{align\_std}, \texttt{consistency},
\texttt{sharp}, \texttt{contrast}, \texttt{colorfulness}\}, with correlation-matrix condition number \condAfter{}
(from \condBefore{}) and eigenvalues from $2.21$ down to $0.34$; standardisation is fitted on the train split and
frozen alongside the basis.

The \nW{} evaluated directions are: six named scenarios (the frozen target re-expressed in this basis, plus
\texttt{ad\_film}, \texttt{teaching\_demo}, \texttt{dynamic\_shot}, \texttt{static\_stable},
\texttt{thumbnail\_cheap}), \nDims{} single-dimension corners, and \nWdir{} draws from Dirichlet($\mathbf{1}$) over
the nine dimensions. Chebyshev scalarisation uses $\min_i w_i(z_i-z_i^{\text{ref}})$ with the reference point
estimated from the same probe rows, which is where part of its disadvantage comes from.

\section{Compute, software and licences}
\label{app:compute}

Everything runs on one node with 8$\times$A800-80GB (driver 535, CUDA 12.8), PyTorch 2.10, \texttt{diffusers} and
\texttt{transformers} in two separate virtual environments because VideoReward pins an older
\texttt{transformers}. Generation and scoring together consumed \gpuHours{} A800-hours; all routing, cold-start and
multi-utility experiments are CPU-bound and take minutes. Model weights are used under their published licences
(Apache-2.0 for CogVideoX and Wan, the LTX-Video and Kandinsky licences, and the Tencent Hunyuan community licence for
the held-out family); prompts come from VBench and VBench-2.0 under their respective research licences, and the human
preference labels from VideoGen-RewardBench.

\section{Reproduction}
\label{app:repro}

The release contains the frozen target and its hash, the evaluator, every method file, the arm registry with exact
checkpoint revisions, per-clip scores and the scripts that emit every number and figure in this paper. The order is:
\texttt{dl\_arms.py} (weights) $\rightarrow$ \texttt{gen.py} (clips) $\rightarrow$ \texttt{score\_arm.py} and
\texttt{score\_feats.py} (scores) $\rightarrow$ \texttt{calib\_target.py} (target, once) $\rightarrow$
\texttt{build\_labels.py} (label matrix) $\rightarrow$ \texttt{gate\_check.py} (\cref{sec:gate}) $\rightarrow$
\texttt{harness/run.py} (routing and cold start) $\rightarrow$ \texttt{pareto\_basis.py} and
\texttt{pareto\_eval.py} (\cref{sec:pareto}). Raw clips total 16\,GB and are not released; the per-clip score tables
are enough to reproduce every table here.

\end{document}

%% file: numbers.tex
\newcommand{\nArms}{16}
\newcommand{\nFamilies}{7}
\newcommand{\costSpan}{121}
\newcommand{\costMin}{0.0025}
\newcommand{\costMax}{0.298}
\newcommand{\nPool}{2{,}029}
\newcommand{\nTrain}{1{,}220}
\newcommand{\nDev}{405}
\newcommand{\nTest}{404}
\newcommand{\nGenPrompt}{899}
\newcommand{\nFullPrompt}{1{,}625}
\newcommand{\nVideos}{64{,}109}
\newcommand{\gpuHours}{2{,}580}
\newcommand{\calibCV}{68.7}
\newcommand{\calibCVsd}{2.6}
\newcommand{\calibBest}{66.2}
\newcommand{\calibVideos}{1{,}705}
\newcommand{\calibPairs}{3{,}060}

\newcommand{\oracleGap}{0.642}
\newcommand{\gapWithin}{0.426}
\newcommand{\gapCross}{0.199}
\newcommand{\gapReal}{47}
\newcommand{\seedSd}{0.429}

\newcommand{\interF}{4.47}
\newcommand{\interP}{1.6\times10^{-63}}
\newcommand{\interVar}{0.585}
\newcommand{\noiseVar}{0.337}

\newcommand{\interRtwoMean}{0.205}
\newcommand{\interRtwoLo}{0.091}
\newcommand{\interRtwoHi}{0.306}
\newcommand{\fullGain}{0.188}
\newcommand{\shufGain}{0.010}

\newcommand{\rbMean}{0.786}

\newcommand{\nDims}{9}
\newcommand{\nDimsRaw}{12}
\newcommand{\condBefore}{282}
\newcommand{\condAfter}{6.5}
\newcommand{\nW}{215}
\newcommand{\nWdir}{200}
\newcommand{\hvOracle}{0.653}
\newcommand{\gainOracle}{0.370}
\newcommand{\gainOracleMin}{0.217}

\newcommand{\gainMulti}{0.031}
\newcommand{\domMulti}{79}

\newcommand{\gainDirect}{0.032}
\newcommand{\domDirect}{78}
\newcommand{\gainCheb}{0.029}
\newcommand{\domCheb}{71}

\newcommand{\bench}{VidRouteBench}
\newcommand{\algo}{OPMU}
\newcommand{\fgEmb}{0.210}
\newcommand{\fgBow}{0.175}
\newcommand{\fgBowDev}{0.009}
\newcommand{\fgEmbDev}{0.168}
\newcommand{\cvMean}{-3.32}
\newcommand{\cvProfile}{+0.04}
\newcommand{\cvRidge}{-2.44}
\newcommand{\cvDim}{-0.23}

\newcommand{\dvProfile}{+0.21}

\newcommand{\dvDim}{+0.25}
\newcommand{\seedSdPanel}{0.42}
\newcommand{\armSpread}{0.64}
\newcommand{\oracleChoice}{0.38}
\newcommand{\bokLo}{0.35}
\newcommand{\bokHi}{0.52}
\newcommand{\weakFour}{+0.001}
\newcommand{\strongOne}{-0.105}
\newcommand{\weakFourCost}{0.296}
\newcommand{\strongOneCost}{0.298}
\newcommand{\sampFront}{0.058}
\newcommand{\sampSave}{23.8}
\newcommand{\sampSaveTop}{31.2}
\newcommand{\gainHistSeed}{0.224}
\newcommand{\gainIndepSeed}{0.155}
\newcommand{\nPanelBok}{333}
\newcommand{\nDevBok}{87}
\newcommand{\rtBO}{0.049}
\newcommand{\rtRone}{0.111}
\newcommand{\rtMone}{0.137}
\newcommand{\rtMthree}{0.151}
\newcommand{\rtMtwo}{0.160}

%% file: tables/tab_arms.tex
\begin{tabular}{llrrrrrr}
\toprule
arm & family & resolution & frames & steps & GPU$\cdot$s & \$/clip & mean $Q$ \\
\midrule
ltx097d\_lr8 & ltx097d & 640$\times$384 & 121 & 8 & 6 & 0.0025 & -1.444 \\
ltx097d\_s8 & ltx097d & 768$\times$512 & 121 & 8 & 10 & 0.0043 & -1.782 \\
ltx097d\_s4 & ltx097d & 768$\times$512 & 121 & 4 & 10 & 0.0043 & -2.074 \\
wan22\_5b\_s20 & wan22\_5b & 832$\times$480 & 121 & 20 & 54 & 0.0227 & -1.355 \\
wan22\_5b\_s10 & wan22\_5b & 832$\times$480 & 121 & 10 & 58 & 0.0243 & -1.814 \\
cog2b\_s25 & cogvideox2b & 720$\times$480 & 49 & 25 & 82 & 0.0344 & -1.485 \\
wan21\_13b\_s25 & wan21\_13b & 832$\times$480 & 81 & 25 & 91 & 0.0381 & -0.985 \\
wan22\_5b\_s40 & wan22\_5b & 832$\times$480 & 121 & 40 & 97 & 0.0405 & -1.120 \\
wan21\_13b\_s12 & wan21\_13b & 832$\times$480 & 81 & 12 & 104 & 0.0433 & -1.650 \\
cog5b\_s25 & cogvideox5b & 720$\times$480 & 49 & 25 & 114 & 0.0474 & -1.403 \\
cog5b\_s15 & cogvideox5b & 720$\times$480 & 49 & 15 & 160 & 0.0665 & -2.063 \\
wan21\_13b\_s50 & wan21\_13b & 832$\times$480 & 81 & 50 & 176 & 0.0735 & -0.624 \\
cog2b\_s50 & cogvideox2b & 720$\times$480 & 49 & 50 & 189 & 0.0787 & -0.726 \\
cog5b\_s50 & cogvideox5b & 720$\times$480 & 49 & 50 & 221 & 0.0920 & -0.501 \\
kandinsky5\_s50 & kandinsky5 & 768$\times$512 & 121 & 50 & 456 & 0.1901 & -0.202 \\
wan22\_a14b\_s40 & wan22\_a14b & 832$\times$480 & 81 & 40 & 714 & 0.2975 & -0.122 \\
\bottomrule
\end{tabular}

%% file: tables/tab_gate.tex
\begin{tabular}{lrl}
\toprule
quantity & value & note \\
\midrule
best single arm (SBM), mean $Q$ & $-0.513$ & \\
per-prompt oracle, mean $Q$ & $+0.129$ & \\
\textbf{oracle gap} & $\mathbf{+0.642}$ & room for any router \\
\midrule
same-seed gap & $+0.426$ & 150 prompts $\times$ 4 arms \\
cross-seed gap & $+0.199$ & pick on seed 0, score on seed 1 \\
\textbf{real fraction} & $\mathbf{47\%}$ & rest is seed noise \\
seed s.d.\ of $Q$ & $0.429$ & same prompt, same arm \\
\midrule
variance: prompt / arm / interaction & 55.3 / 11.3 / 33.4\% & \\
interaction $F(447,600)$ & $4.47$ & $p=1.6\times10^{-63}$ \\
variance components: interaction / noise & $0.585$ / $0.337$ & interaction is 63\% \\
\midrule
out-of-fold $R^2$ of interaction & $0.091$--$0.306$ & mean $0.205$, per arm \\
routing gain, full labels & $+0.188$ & \\
routing gain, shuffled labels & $+0.010\pm0.012$ & control \\
\bottomrule
\end{tabular}

%% file: tables/tab_router.tex
\begin{tabular}{llrr}
\toprule
prompt representation & router & full-label gain & $\Delta$ vs.\ embedding \\
\midrule
\multicolumn{4}{l}{\emph{5-fold CV over train$\cup$dev}} \\
TF-IDF bag-of-words (1--2 grams) & ridge & $+0.175$ & $-17\%$ \\
Qwen3-Embedding-0.6B & ridge & $\mathbf{+0.210}$ & -- \\
\multicolumn{4}{l}{\emph{train $\rightarrow$ dev, the low-data regime}} \\
TF-IDF bag-of-words (1--2 grams) & ridge & $+0.009$ & $-95\%$ \\
Qwen3-Embedding-0.6B & ridge & $\mathbf{+0.168}$ & -- \\
\midrule
\multicolumn{4}{l}{\emph{references}} \\
shuffled labels & ridge & $+0.010\pm0.012$ & control \\
per-prompt oracle & -- & $+0.642$ & upper bound \\
\bottomrule
\end{tabular}

%% file: tables/tab_cold.tex
\begin{tabular}{llrrr}
\toprule
 & & \multicolumn{2}{c}{recovery} & \\
\cmidrule(lr){3-4}
method & what the probe gives it & 5-fold & dev & probe cost (\$) \\
\midrule
\textsf{mean} (M0) & one scalar average & $-3.32$ & $-0.66$ & $1.18$ / $2.36$ / $5.90$ \\
\textsf{ridge}$_K$ (M2) & $K$ (prompt, quality) pairs & $-2.44$ & $-0.52$ & same \\
\textsf{profile} (M1) & quality vs.\ known arms & $\mathbf{+0.04}$ & $+0.21$ & same \\
\textsf{dim-profile} (ours, rejected) & per-dimension vs.\ known arms & $-0.23$ & $+0.25$ & same \\
\bottomrule
\end{tabular}

%% file: tables/tab_pareto.tex
\begin{tabular}{llrrrr}
\toprule
utilities & rule & gain & min gain & dominance & $\Delta$HV \\
\midrule
oracle utilities (upper bound) & linear & $+0.370$ & $+0.217$ & $\mathbf{100\%}$ & $+0.653$ \\
oracle utilities & Chebyshev & $+0.361$ & $+0.209$ & $\mathbf{100\%}$ & $+0.617$ \\
\midrule
\algo{} (multi-output, ours) & linear & $+0.031$ & $+0.000$ & $\mathbf{79\%}$ & $+0.081$ \\
per-scenario router & linear & $\mathbf{+0.032}$ & $+0.000$ & $78\%$ & $+0.082$ \\
\algo{} & Chebyshev & $+0.029$ & $+0.000$ & $71\%$ & $+0.069$ \\
\bottomrule
\end{tabular}

%% file: tables/tab_samples.tex
\begin{tabular}{llrrrr}
\toprule
policy & reads prompt & actions & gain vs.\ B0 & win rate & loses to B1 ($B{=}5$) \\
\midrule
B0 prompt-agnostic hull & no & $(a,1)$ & $0$ & $50.0\%$ & -- \\
B1 prompt-agnostic resampling & no & $(a,k)$ & $+0.049$ & $51.2\%$ & -- \\
R1 router, one sample & yes & $(a,1)$ & $+0.111$ & $52.8\%$ & $46\%$ \\
R2-M1 router $+$ per-arm $g_k$ & yes & $(a,k)$ & $+0.137$ & $53.4\%$ & $43\%$ \\
R2-M3 router $+$ predicted variance & yes & $(a,k)$ & $+0.151$ & $53.8\%$ & $41\%$ \\
R2-M2 router, best-of-$k$ targets (ours) & yes & $(a,k)$ & $\mathbf{+0.160}$ & $\mathbf{54.0\%}$ & $\mathbf{40\%}$ \\
\midrule
oracle (true values) & -- & $(a,k)$ & $+0.793$ & $68.9\%$ & $0\%$ \\
\bottomrule
\end{tabular}